\documentclass[letterpaper]{article} 
\usepackage{aaai2026}  
\usepackage{times}  
\usepackage{helvet}  
\usepackage{courier}  
\usepackage[hyphens]{url}  
\usepackage{graphicx} 
\usepackage{natbib}  
\usepackage{caption} 
\usepackage{algorithm}
\usepackage{algorithmic}

\usepackage{multirow}
\usepackage{booktabs}
\usepackage{graphicx}
\usepackage{colortbl}
\usepackage{diagbox}
\usepackage{amsmath}
\usepackage{bbding}
\usepackage{pifont}
\usepackage{enumitem}

\usepackage{newfloat}
\usepackage{listings}
\DeclareCaptionStyle{ruled}{labelfont=normalfont,labelsep=colon,strut=off} 
\floatstyle{ruled}
\newfloat{listing}{tb}{lst}{}
\floatname{listing}{Listing}
\title{Asymmetric Cross-Modal Knowledge Distillation: Bridging Modalities with Weak Semantic Consistency}
\author{
    Riling Wei\textsuperscript{\rm 1}\equalcontrib,
    Kelu Yao\textsuperscript{\rm 1}\equalcontrib,
    Chuanguang Yang\textsuperscript{\rm 2},
    Jin Wang\textsuperscript{\rm 3},
    Zhuoyan Gao\textsuperscript{\rm 1},
    Chao Li\textsuperscript{\rm 1}\thanks{Corresponding author.}
}

\affiliations{
    \textsuperscript{\rm 1}Research Center for Space Computing System , Zhejiang Laboratory, Hangzhou, China\\
    \textsuperscript{\rm 2}Institute of Computing Technology, Chinese Academy of Sciences, Beijing, China\\
    \textsuperscript{\rm 3}The University of Hong Kong, Hong Kong, China\\
    \{weirl, yaokelu, gaozy, lichao\}@zhejianglab.org, yangchuanguang@ict.ac.cn, wj0529@connect.hku.hk
}

\usepackage{bibentry}

\begin{document}

\maketitle

\begin{abstract}
Cross-modal Knowledge Distillation has demonstrated promising performance on paired modalities with strong semantic connections, referred to as Symmetric Cross-modal Knowledge Distillation (SCKD). However, implementing SCKD becomes exceedingly constrained in real-world scenarios due to the limited availability of paired modalities. To this end, we investigate a general and effective knowledge learning concept under weak semantic consistency, dubbed Asymmetric Cross-modal Knowledge Distillation (ACKD), aiming to bridge modalities with limited semantic overlap. Nevertheless, the shift from strong to weak semantic consistency improves flexibility but exacerbates challenges in knowledge transmission costs, which we rigorously verified based on optimal transport theory. To mitigate the issue, we further propose a framework, namely SemBridge, integrating a Student-Friendly Matching module and a Semantic-aware Knowledge Alignment module. The former leverages self-supervised learning to acquire semantic-based knowledge and provide personalized instruction for each student sample by dynamically selecting the relevant teacher samples. The latter seeks the optimal transport path by employing Lagrangian optimization. To facilitate the research, we curate a benchmark dataset derived from two modalities, namely Multi-Spectral (MS) and asymmetric RGB images, tailored for remote sensing scene classification. Comprehensive experiments exhibit that our framework achieves state-of-the-art performance compared with 7 existing approaches on 6 different model architectures across various datasets.
\end{abstract}

\begin{links}
    \link{Code}{https://github.com/weirl-922/ACKD}
\end{links}

\section{Introduction}

Cross-modal Knowledge Distillation (CMKD) \cite{huo2024c2kd, wang2024distilvpr,dai2021learning,li2022cross, xue2022modality} has demonstrated remarkable performance in various tasks such as visual recognition  \cite{zhao2024simdistill, lu2024breaking,kim2024labeldistill} and audio-visual classification \cite{sarkar2024xkd,huo2024c2kd,ren2021learning}, by transferring complementary knowledge across modalities from teacher to student models.
Compared to conventional computer vision tasks, remote sensing (RS) tasks often involve richer and more diverse data modalities, e.g., Multi-Spectral (MS) images~\cite{kettig1976classification}, Hyper-Spectral (HS) images~\cite{landgrebe2002hyperspectral}, Light Detection And Ranging (LiDAR)~\cite{reutebuch2005light}, etc., making them particularly well-suited for cross-modal learning.
\begin{figure}[t]
	\centering
	\setlength{\belowcaptionskip}{0.cm}
	\includegraphics[width=8.5 cm]{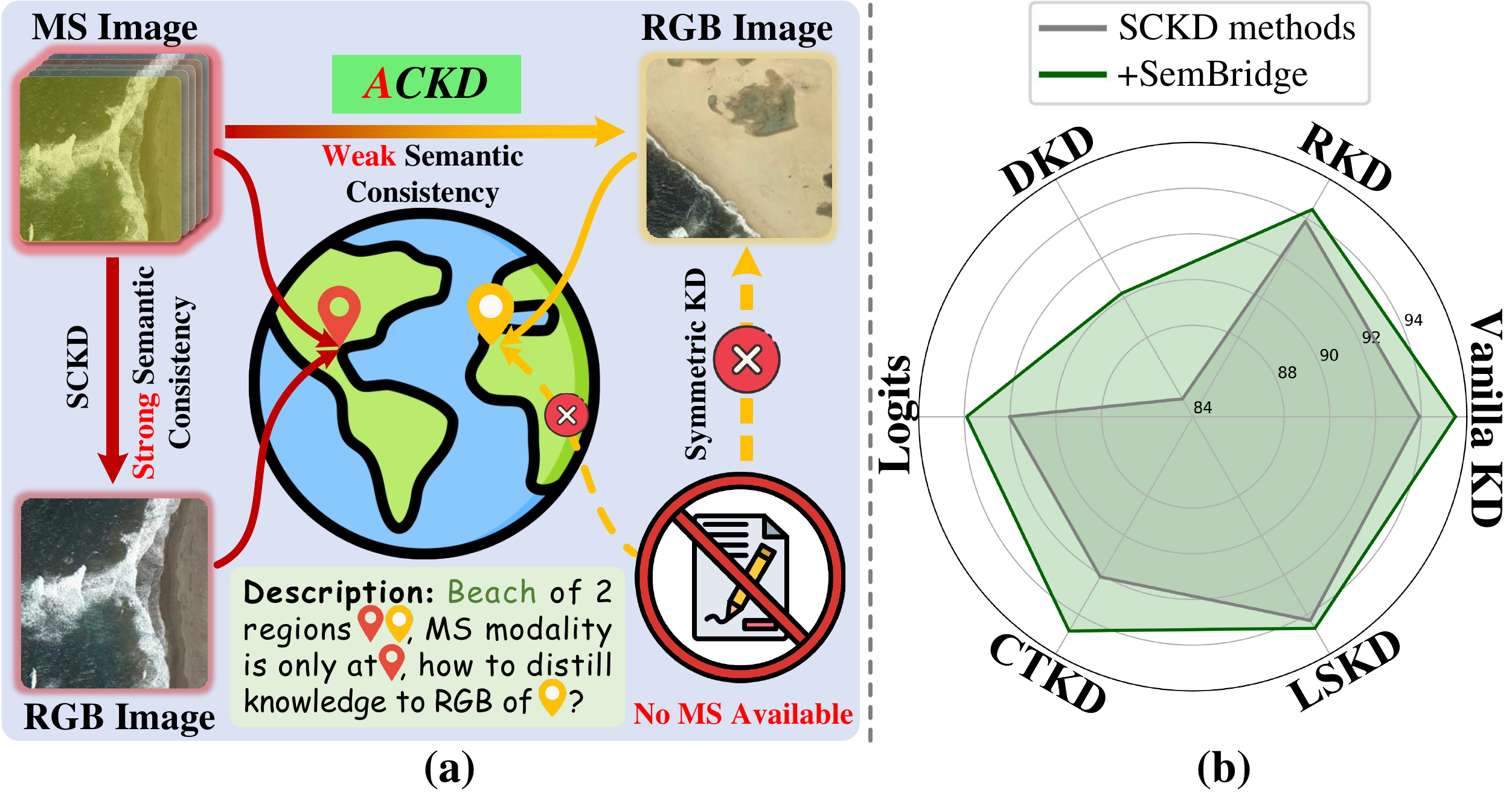}
    \caption{(a) SCKD distills knowledge between modalities from the same location, assuming strict semantic alignment. In contrast, ACKD relaxes this constraint, enabling cross-modal transfer with only weak semantic consistency, regardless of location. This allows a small MS dataset to benefit a larger RGB set. (b) The proposed SemBridge further boosts the performance of SCKD approaches (DKD, RKD, Vanilla KD, LSKD, CTKD, and Logits) under ACKD settings.}
	\label{fig1}
\end{figure}
In recent years, this potential has attracted increasing attention, and numerous studies have explored the application of CMKD to remote sensing scenarios.
In scene classification~\cite{ChengGong2}, researchers~\cite{LiuH, ShinHK} have used MS images as the teacher modality to distill knowledge into RGB images via CMKD, resulting in significantly improved performance of the RGB-based models. 
In land cover classification~\cite{phiri2017developments}, Wang \textit{et al.}~\cite{wang2023cross} applied CMKD to address the issue of missing modalities during inference, and demonstrated its effectiveness across several multi-modal RS datasets.
Despite its promising potential in RS, CMKD still faces notable challenges in real-world applications.
Most existing approaches~\cite{LiuH, ShinHK, wang2023cross} inherently assume that the modalities used by the teacher and student share the same semantic content (\textit{i.e.,} paired data), a setup we collectively refer to as Symmetric Cross-modal Knowledge Distillation (SCKD).

\begin{figure}[t]
	\centering
	\setlength{\belowcaptionskip}{0.cm}
	\includegraphics[width=8.5cm]{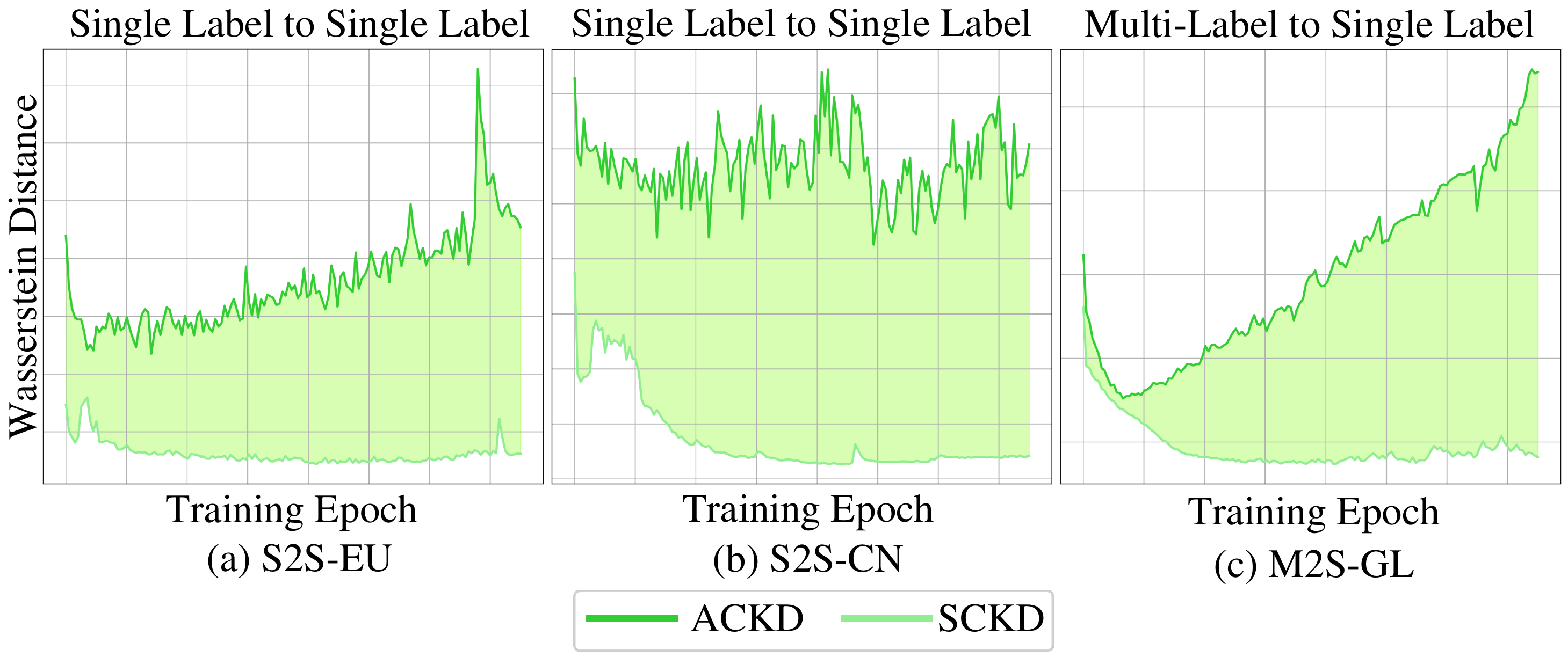}
    \caption{Wasserstein distance between ACKD and SCKD on three datasets. ACKD consistently incurs higher transport costs than SCKD during training, reflecting the challenge of cross-modal alignment in asymmetric settings.}
	\label{fig2}
\end{figure}

However, in practice, the application of SCKD is often constrained by the scarcity of paired data, primarily due to the quantity imbalance arising from the high acquisition cost of teacher modalities.
For example, MS images, which are commonly employed as teacher modalities, typically outperform RGB images in scene-understanding tasks due to their higher spectral resolution~\cite{park2007multispectral, ma2023multispectral}. Nevertheless, collecting MS data requires specialized equipment, posing significant challenges for large-scale deployment. In contrast, lower-information-density modalities, such as RGB images, are far more accessible through satellites, UAVs, and other widely available platforms~\cite{LiuH, ShinHK, ChengGong2}.
As a result, only a small fraction of RGB images are accompanied by corresponding MS modalities, limiting the scalability and practicality of SCKD.

This challenge underscores the need for more flexible distillation strategies that can operate effectively under unpaired or weakly paired settings.
A natural and important question thus arises: \textit{Is it possible to distill knowledge between modalities that do not share strong semantic correspondence, such as MS images collected from Europe and RGB images captured in Asia?}
We refer to this setting as Asymmetric Cross-modal Knowledge Distillation (ACKD), as illustrated in Figure~\ref{fig1}.

Accordingly, ACKD is proposed to overcome the limitations of SCKD in unpaired scenarios by facilitating knowledge transfer between modalities with significant semantic discrepancies.
As shown in Table~\ref{tab1} and Table~\ref{tab2}, multiple state-of-the-art knowledge distillation methods fail to achieve satisfactory performance when applied directly to ACKD. 
In certain cases, the performance even drops below that of the uni-modal baseline without any distillation, indicating that directly transferring SCKD strategies to asymmetric scenarios is ineffective.

To this end, we conducted a theoretical analysis grounded in optimal transport theory~\cite{santambrogio2015optimal} and demonstrated that the key bottleneck of ACKD lies in its inherently higher cost of knowledge transfer. 
Compared to SCKD, the substantial semantic gap between input modalities leads to significantly increased transport costs during training.
To further support this observation, we visualize the Wasserstein distance~\cite{rubner2000earth} in Figure~\ref{fig2}, a widely used metric in optimal transport theory~\cite{solomon2015convolutional,chen2020graph,frogner2015learning} that quantifies the cost of knowledge transfer across modalities. 
The results clearly show that ACKD incurs a much higher transport cost than SCKD.
Further analysis in both the label space and latent space reveals that weak semantic consistency not only increases the transport cost but also reduces mutual information~\cite{batina2011mutual} between modalities, thereby diminishing the overlap of transferable knowledge between the teacher and the student.
These findings highlight the urgent need for dedicated distillation frameworks tailored to ACKD, capable of bridging the semantic gap and enhancing cross-modal knowledge alignment.

To tackle the aforementioned challenges in ACKD, we propose SemBridge, a novel distillation framework designed to optimize knowledge transfer under semantic misalignment.
Specifically, SemBridge integrates two plug-and-play modules: the Student-Friendly Matching (SFM) module and the Semantic-aware Knowledge Alignment (SKA) module.
The SFM module aims to reduce transport costs by adaptively establishing suitable teacher-student matching. 
Inspired by the strong semantic correspondence typically assumed in SCKD, SFM first assigns an initial teacher to each student sample based on semantic similarity. 
Moreover, drawing inspiration from the human educational paradigm, SFM enables student samples to dynamically select their subsequent teachers throughout training based on evolving learning needs.
In parallel, the SKA module is introduced to further optimize the transport process. 
It first formulates an intra-modal transport plan via Lagrangian optimization, capturing semantic structure within each modality. 
Based on this, cross-modal transport plans are constructed separately for both the teacher and student modalities, facilitating more efficient and semantically aligned knowledge transfer.

Moreover, to facilitate our research, we construct a dataset benchmark with 3 sub-datasets, including MS images and asymmetric RGB images, namely S2S-EU, S2S-CN, and M2S-GL, respectively. The dataset includes a total of 70,414 MS images and 63,549 unpaired RGB images across diverse scene categories on Earth. To evaluate the generalization capability of SemBridge, we select MS images collected by different equipment with various numbers of spectral channels.

In our experiments, we evaluate SemBridge under both homogeneous and heterogeneous model architectures by distilling knowledge from both multi-label and single-label teachers into single-label students.
The results show that SemBridge not only enables Vanilla KD~\cite{hinton2015distilling} to achieve state-of-the-art performance among seven baseline methods but also consistently improves the performance of other SCKD-based approaches.

Our contribution can be summarized as:

\begin{itemize}
[noitemsep,nolistsep,,topsep=0pt,parsep=0pt,partopsep=0pt]
\item To the best of our knowledge, we are the first to explore Asymmetric Cross-modal Knowledge Distillation (ACKD), a promising concept that broadens the application scope of Symmetric Cross-modal Knowledge Distillation (SCKD).
\item We propose SemBridge, a plug-and-play framework including Student-Friendly Matching and Semantic-aware Knowledge Alignment, that enables existing SCKD methods to achieve significant performance gains in ACKD by explicitly optimizing the transport cost.
\item We construct a new benchmark consisting of three sub-datasets with MS and asymmetric RGB image pairs to facilitate evaluation under real-world asymmetry.
\end{itemize}

\begin{table}[t]
\centering
\begin{tabular}{ll}
\toprule
\textbf{Symbol} & \textbf{Description} \\
\midrule
$\mathcal{D}$, $\mathcal{D}_{match}$ & Unpaired and Matched dataset \\
$V, G, \tilde{G}$ & MS, RGB and Psedo-RGB modality \\
$\mathcal{T}, \mathcal{S}$ & Teacher and student \\
$f_T, f_S$ & Feature extractors \\
$h_T, h_S$ & Classifiers \\
$\mathcal{M}$ & Matcher \\
$\mathcal{M}_V$, $\mathcal{M}_G$ & MS and RGB Encoder of $\mathcal{M}$\\
$z_T, z_S$ & Unfused features \\
$\overline{z}_T, \overline{z}_S$ & Fused features \\
$\textbf{p}_T, \textbf{p}_S$ & Outputs logits \\
$v, \tilde{g}$ & Representation of MS and Psedo-RGB\\
Planner & the proposed Planner \\
\bottomrule
\end{tabular}
\caption{Summary of Notations}
\label{tab1}
\end{table}

\section{Related Works}
\textbf{Remote Sensing (RS) Scene Classification} aims to categorize geographic areas based on their semantic content. Early approaches relied on handcrafted features extracted from RGB images~\cite{Anil_c,ZhangYS}. Recently, deep learning methods have achieved notable success due to the strong generalization ability of neural networks~\cite{ZouQ,ChengG}. However, in complex scenes, simply increasing network width or depth does not always improve performance, as RGB images often suffer from low information density. To address this, multispectral (MS) images have been introduced~\cite{GomezP}, offering richer information via additional spectral bands. While MS images generally outperform RGB ones, their acquisition requires specialized sensors, and the increased spectral channels lead to higher computational costs. To alleviate these issues, researchers~\cite{LiuH,ShinHK} have proposed cross-modal knowledge distillation (CMKD) to transfer semantic knowledge from MS to RGB images, enabling efficient inference using only the RGB modality.

\textbf{Symmetric modality-based Knowledge Distillation}. Knowledge distillation (KD) was first proposed by Hinton \textit{et al.}~\cite{hinton2015distilling} for optimizing the computational cost and memory consumptions on devices with limited computation or storage resources, which is regarded as uni-modality-based KD as both the teacher and student take the same modality as input. KD can be categorized into response-based~\cite{sun2024logit,zhao2022decoupled,ba2014deep,li2023curriculum,hao2023ofa,hinton2015distilling}, feature-based~\cite{yang2024vitkd, yang2024clip} as well as relation-based~\cite{park2019relational, yang2022cross, yang2023online} KD determined by which parts of the model are distilled. 
With the remarkable success of cross-modal learning~\cite{kaur2021comparative,HeCMR}, symmetric cross-modal knowledge distillation (SCKD) has gathered much attention, aiming to conduct knowledge from discriminate modalities to the weaker ones. 

\section{Methodology}
\subsection{Overview}
Given a teacher model \(\mathcal{T}\) and a student model \(\mathcal{S}\) taking MS modality \(V\), and RGB modality \(G\) respectively. The dataset \(\mathcal{D}\) contains unpaired samples from \(C\) classes:
$\mathcal{D} = \{ (V_k^c)_{k=1}^{K_c}, (G_n^c)_{n=1}^{N_c} \}_{c=1}^C,$
where $c$ denotes the class index, and $K_c$, $N_c$ are the number of MS and RGB samples in the $c$-th class, respectively. Let $f_T$ and $f_S$ denote the feature extractors of \(\mathcal{T}\) and \(\mathcal{S}\), and $h_T$, $h_S$ be their respective classifiers. Let $z_T = f_T(V)$,\quad $z_S = f_S(G)$ be unfused features, and $\overline{z}_T$, $\overline{z}_S$ be fused feature representations, obtained by applying adaptive average pooling on $z_T$ and $z_S$. $\textbf{p}_T = h_T(\overline{z}_T)$,\quad $\textbf{p}_S = h_S(\overline{z}_S)$ are output logits.
We design a matcher $\mathcal{M} = (\mathcal{M}_V, \mathcal{M}_G)$, which consisted of 2 encoders $\mathcal{M}_V$ and $\mathcal{M}_G$ to project $V$ and pseudo-RGB images $\tilde{G}$ to corresponding representations $v = \mathcal{M}_V(V), \tilde{g} = \mathcal{M}_G(\tilde{G}).$
\begin{figure}[t]
	\centering
	\setlength{\belowcaptionskip}{0.cm}
	\includegraphics[width=8.4 cm]{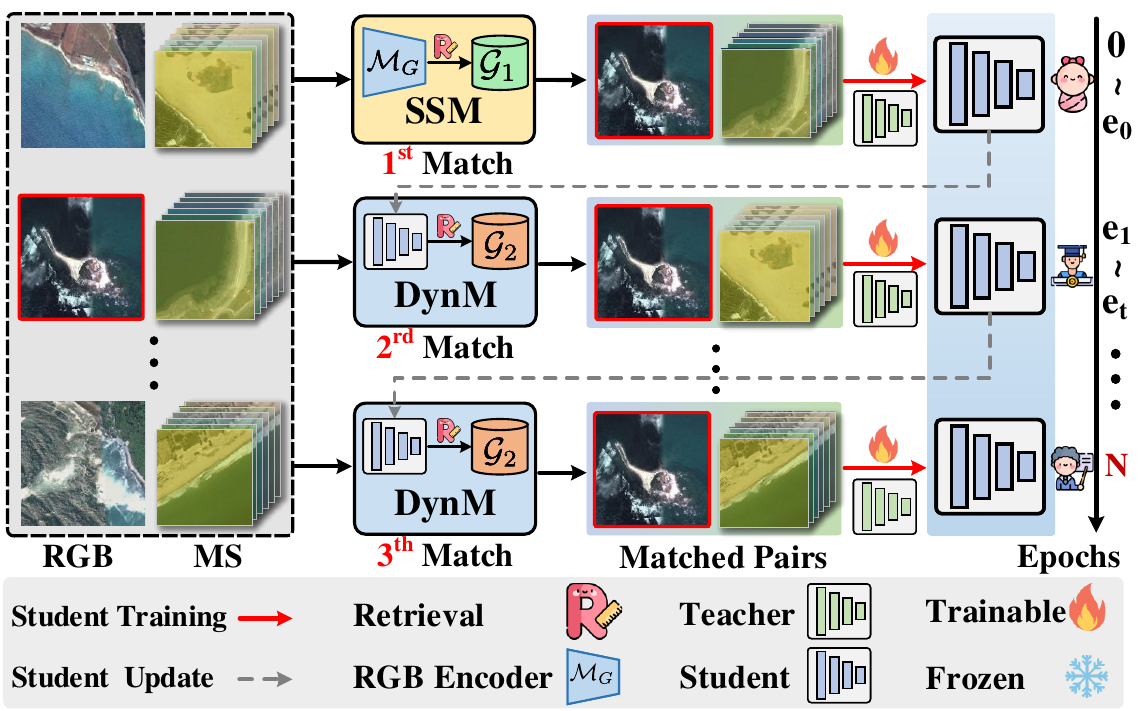}
    \caption{Illustration of the proposed student friendly matching strategy consisting of SSM for the first matching and Dyn. M allowing student select proper teacher samples dynamically at the training stage. In DynM, current student model is designed to involved.}
	\label{fig3}
\end{figure}

$\mathcal{T}$ is first trained in an offline manner. During this time, a matcher $\mathcal{M}$ is also trained for the initial matching, which will be introduced in the next sections.

At the training stage of $\mathcal{S}$, we propose SemBridge consisting of a Student-Friendly Matching (SFM) module and a Semantic-aware Knowledge Alignment (SKA) module to select a teacher-student sample with greater semantic consistency and updated by the current student model several times. Finally, we finalize an optimal transport plan for weak semantic consistency modalities via the SKA module. 
\begin{table}[t]
	\setlength{\tabcolsep}{7.5pt}
	\renewcommand{\arraystretch}{1.0}
\centering
\begin{tabular}{lccc}
\toprule
\textbf{Subset} & \textbf{S2S-EU} & \textbf{S2S-CN} & \textbf{M2S-GL} \\
\midrule
MS Label & Single & Single & Multiple\\
RGB Label & Single & Single & Single\\
Devices & Sentinel-2 & Tiangong-2 & Sentinel-2 \\
MS bands & 10 & 14 & 10 \\
Resolution & $64 \times 64$ & $128 \times 128$ & $120 \times 120$\\
Categories & 10 & 10 & 15 \\
\bottomrule
\end{tabular}
\caption{Details of the proposed dataset benchmark.}
\label{tab2}
\end{table}
\subsection{Optimal transport analysis}
To demonstrate the knowledge transport cost caused by weak semantic consistency, we utilize the Wasserstein distance to compare the output logits extracted from strong (SCKD) and weak (ACKD) semantic consistency, respectively. Wasserstein distance is a common tool to measure optimal transport, which is used to evaluate the difficulty of knowledge propagation.
Suppose $x_S$ and $x_T$ are the inputs of two modalities, the corresponding probability distributions: $f_S(x_S) \sim \mathcal{P}_S$ and $f_T(x_T) \sim \mathcal{P}_T$. The Wasserstein distance $\mathcal{W}$ can be formulated as:
{\small
\begin{equation}
\label {eq1}
\ \mathcal{W}(\mathcal{P}_S, \mathcal{P}_T) = \inf_{\pi \in \Pi(\mathcal{P}_S, \mathcal{P}_T)} {E}_{(x_S, x_T) \sim \pi} \left[ c(f_S(x_S), f_T(x_T)) \right],
\end {equation}
}
where $c(\cdot)$ is the distance measurement. $f_S$ and $f_T$ are the feature extractors. $\Pi(\mathcal{P}_S, \mathcal{P}_T)$ is a set of candidate point of $\mathcal{P}_S(x_S)$ and $\mathcal{P}_T(x_T)$. $\Pi(\mathcal{P}_S, \mathcal{P}_T)$ is a joint distribution satisfying $\int \pi(x_S, x_T) \, dx_T = \mathcal{P}_S(x_S), \quad \int \pi(x_S, x_T) \, dx_S = \mathcal{P}_T(x_T)$. 

As shown in Figure~\ref{fig2}, due to significant cost, ACKD becomes more challenging compared to SCKD. Therefore, we are committed to finding a reasonable transport plan $\pi$ to optimize the cost of knowledge propagation.

\subsection{Student-Friendly Matching (SFM)}
The first step to optimize the cost is to select a suitable teacher sample for each student with greater semantic consistency motivated by SCKD. Then, inspired by human educational wisdom, dynamic matching is proposed to select different teacher samples for students during their learning period, as shown in Figure~\ref{fig3}. To be specific, by matching reasonable teacher samples for students, the cost can be optimized as $\pi(i) = \arg \min_j \| f_S(x_S^i) - f_T(x_T^j) \|^2$. In other words, an optimal joint distribution is found by selecting teacher samples as Equ.~(\ref{eq2}):
{
\small
\begin{equation}
    \label{eq2}
\ \pi^*(x_S, x_T) = 
\begin{cases} 
1, & \text{if } x_T = \arg \min_{x_T'} \| f_S(x_S) - f_T(x_T') \|^2 \\
0, & \text{otherwise}
\end{cases}
\end {equation}
}

Subsequently, the optimized $\mathcal{W}$ can be implemented as:
{
\small
\begin{equation}
\label{eq3}
\ \mathcal{W}(P_S, P_T)=\sum_i \pi^*(x_S^i, x_T^j) ||f_S(x_S^i)-f_T(x_T^j)||^2.
\end {equation}
\subsubsection{Construct retrieval teacher galleries.} Before training $\mathcal{S}$, we construct two teacher galleries, noted as $\mathcal{G}_1$ and $\mathcal{G}_2$, to enable the student to retrieve teacher samples in SFM. The trained $\mathcal{T}$ and the MS encoder $\mathcal{M}_V$ are used to project MS samples into embeddings and logits and saved in $\mathcal{G}_1$ and $\mathcal{G}_2$ respectively. Specifically, for the $c$-th class, let: $V^c:\{V_k^c \mid k=1, \dots, K_c\}$ denote the MS samples. $\mathcal{M}_V$ and $\mathcal{T}$ project it into:
\begin{equation}
\label{eq4}
    v_k^c = \mathcal{M}_V(V_k^c), \quad \textbf{p}_{T,k}^c = \mathcal{T}(V_k^c), \quad \forall k = 1, \dots, K_c,
\end{equation}
and save $v_k^c$ and $\textbf{p}_{T,k}^c$ in $\mathcal{G}_1$ and $\mathcal{G}_2$ respectively.
\subsubsection{Self-supervised Semantic-aware Matching (SSM).}
To learn a semantic-aware matcher without relying on paired RGB images, we utilize only $V$ from the unpaired dataset $\mathcal{D}$. Specifically, we extract the RGB bands from each $V$ to construct a pseudo-RGB image $\tilde{G}$. Since $V$ and $\tilde{G}$ originate from the same source, they naturally share the same semantic content and are treated as positive pairs for self-supervised learning~\cite{jing2020self}.
It should be noted that the split channels are determined by the modality of the student model. For example, in this task, the student modality is the RGB image. Hence, we just split R, G, and B channels from $V$. 

Then, InfoNCE loss~\cite{gutmann2010noise} is employed to optimize the matcher to learn the semantic difference in Contrastive Language–Image Pretraining (CLIP)-based manner~\cite{radford2021learning}.
\begin{equation}
    \mathcal{L}_{V \rightarrow G} = -\log \frac{\exp(v \cdot \tilde{g}^+)}{\sum_{b=1}^{N} \exp(v \cdot \tilde{g}_b)},
\end{equation}
\begin{equation}
    \mathcal{L}_{G \rightarrow V} = -\log \frac{\exp(\tilde{g} \cdot v^+)}{\sum_{b=1}^{N} \exp(\tilde{g} \cdot v_b)},
\end{equation}
where $N$ is the batch size. $\tilde{g}^+$ and ${v}^+$ are positive samples. The total semantic-aware contrastive loss is defined as:
\begin{equation}
    \mathcal{L}_{\text{semantic}} = \frac{1}{2} \left( \mathcal{L}_{V \rightarrow G} + \mathcal{L}_{G \rightarrow V} \right).
\end{equation}

After training $\mathcal{M}$, we use it to select the most semantically consistent teacher samples for each student sample within the same class. For the $c$-th class, let: $G^c:\{G_n^c \mid n=1, \dots, N_c\}$ denote RGB samples and projected into $ g_n^c = \mathcal{M}_G(G_n^c)$. Then, for each $g_n^c$, we compute its cosine similarity $\textbf{cos}(\cdot, \cdot)$ with all teacher embeddings $v_k^c$ from $\mathcal{G}_1$ to form a similarity matrix:
\begin{equation}
\label{eq5}
    \Phi_n^c = \left[ \textbf{cos}(g_n^c, v_1^c), \textbf{cos}(g_n^c, v_2^c), \dots, \textbf{cos}(g_n^c, v_{K_c}^c) \right].
\end{equation}

By stacking all similarity vectors, we obtain the class-wise similarity matrix:
{\small
\begin{equation}
\label{eq6}
    \Phi^c = 
    \begin{bmatrix}
        \Phi_1^c \\
        \Phi_2^c \\
        \vdots \\
        \Phi_{N_c}^c
    \end{bmatrix}
    \in R^{N_c \times K_c}.
\end{equation}
}

Next, for each student sample $G_n^c$ with embedding $g_n^c$, we select the teacher sample with the highest semantic similarity: $\textbf{k}^* = \arg\max_{\textbf{k}} \Phi_{n,\textbf{k}}^c.$
This yields the matched sample pairs for the $c$-th class:
\begin{equation}
\label{eq7}
    \mathcal{D}_{\text{match}}^c = \left\{ \left( V_{\textbf{k}^*}^c, G_n^c \right) \mid n = 1, \dots, N_c \right\}.
\end{equation}
\subsubsection{Dynamic Matching (DynM).} Inspired by human education systems where students are guided by different teachers throughout their learning journey, we propose a DynM strategy. Instead of relying on a fixed teacher, DynM periodically updates the matched teacher-student pairs during training. This allows the student to absorb knowledge from multiple teacher samples, thereby reducing semantic bias and improving generalization.

First, we compute the output logits of the student $\textbf{p}_{S,n}^{c} = \mathcal{S}(G_n^c)$ from the $c$-th class. 
Then, we calculate the Kullback-Leibler (KL) divergence with temprature $\gamma$ between the $n^{th}$ student prediction $\textbf{p}_{S,n}^c$ and all candidate teacher samples $\textbf{p}_{T,k}^c$ from $\mathcal{G}_2$:
{
\small
\begin{equation}
\label{eq8}
    \Omega_n^c=\left[ 
        \mathrm{KL}(\textbf{p}_{S,n}^c \| \textbf{p}_{T,1}^c; \gamma),\ 
        \dots,\ 
        \mathrm{KL}(\textbf{p}_{S,n}^c \| \textbf{}_{T, K_c}^c; \gamma)
    \right].
\end{equation}
}
Unlike selecting teacher samples with maximum semantic similarity to acquire basic knowledge in the early stage of learning at SSM, DynM encourages the student to select more challenging samples, facilitating a progressive transition from easy to difficult knowledge:
\begin{equation}
\label{eq9}
    \textbf{k}^* = \arg\min_\textbf{k} \Omega_{n,\textbf{k}}^c.
\end{equation}
The time of selecting new teachers in human educational systems almost depends on the years of study in the current stage. Based on this, DynM is performed several times along the learning journey as shown in Figure~\ref{fig3}. Inspired by curriculum learning~\cite{bengio2009curriculum}, the time of per matching is gradually extended with the increment of knowledge diversity and implemented as:
\begin{equation}
	\label{eq10}
	\ e_t = e_0 + \sum_{i=1}^{t} (\Delta e + e_{\mu} (i-1)).
\end {equation}
Here, $t$ is the number of times to perform DynM. When $t=1$, the initial DynM is started and $e_{0}$ is the initial matching time (epoch). 
\begin{table}[t]
	\setlength{\tabcolsep}{3.5pt}
	\renewcommand{\arraystretch}{0.9}
	\begin{center}
		\begin{tabular}{l| c|| c c ||c c|| c c}
			\toprule
			&\multirow{2}{*}{\textbf{Model}} & \multicolumn{2}{c||}{\textbf{S2S-EU}} & \multicolumn{2}{c||}{\textbf{S2S-CN}} & \multicolumn{2}{c}{\textbf{M2S-GL}}\\
			\cmidrule(r){3-4} \cmidrule(r){5-6} \cmidrule(r){7-8}
			&&\textbf{OA} & \textbf{F1} &\textbf{OA} & \textbf{F1} &\textbf{OA} & \textbf{F1}\\
			\midrule
			\multirow{9}{*}{\rotatebox[origin=c]{90}{\textbf{Homogeneous model}}}&T:ResNet34 & 95.3 & 95.1 & 96.8 & 97.0 & / & 80.8\\
			&S:ResNet34 & 91.7 & 91.6 & 94.9 & 94.4 & 94.9 & 93.2\\
            & +SemBridge & \textbf{93.7} & \textbf{93.6} & \textbf{96.2} & \textbf{95.8} & \textbf{96.6} & \textbf{95.1} \\
            
			\cmidrule{2-8}
			&T:MobileNetV2 & 95.2 & 95.0 & 95.3 & 95.5 & / & 75.2 \\
			&S:MobileNetV2 & 89.4 & 89.1 & 92.3 & 91.3 & 92.9 & 90.3\\
			& +SemBridge & \textbf{91.7} & \textbf{91.5} & \textbf{93.6} & \textbf{92.8} & \textbf{93.9} & \textbf{91.7}\\
			\cmidrule{2-8}
			&T:ShuffleNetV2 & 92.3 & 92.0 & 93.7 & 93.5 & / & 70.3\\
			&S:ShuffleNetV2 & 85.9 & 85.6 & 90.0 & 88.8 & 88.8 & 85.5\\
			&+SemBridge & \textbf{88.4} & \textbf{88.1} & \textbf{91.4} & \textbf{90.6} & \textbf{90.8} & \textbf{87.8}\\
			\midrule
			\midrule
			\multirow{9}{*}{\rotatebox[origin=c]{90}{\textbf{Heterogeneous model}}}
			&T:ResNet34 & 95.3 & 95.1 & 96.8 & 97.0 & / & 80.8\\
			&S:MobileNet & 89.4 & 89.1 & 92.3 & 91.3 & 92.9 & 90.3\\
			&+SemBridge &\textbf{92.1} & \textbf{91.9} & \textbf{93.5} & \textbf{92.9} & \textbf{93.9} & \textbf{91.7}\\
			\cmidrule{2-8}
			&T:ResNet34 & 95.3 & 95.3 & 96.8 & 96.8 & / & 85.3 \\
			&S:ShuffleNetV2 & 85.9 & 85.6 & 90.0 & 88.8 & 88.8 & 85.5 \\
			&+SemBridge & \textbf{87.9} & \textbf{87.6} & \textbf{91.0} & \textbf{89.8} & \textbf{90.3} & \textbf{87.8}\\
			\cmidrule{2-8}
			&T:MobileNetV2 & 95.2 & 95.0 & 95.3 & 95.5 & / & 75.2\\
			&S:ShuffleNetV2 & 85.9 & 85.6 & 90.0 & 88.8 & 88.8 & 85.5\\
			&+SemBridge & \textbf{87.8} & \textbf{87.4} & \textbf{91.6} & \textbf{90.7} & \textbf{89.7} & \textbf{87.5}\\
			\hline
		\end{tabular}
        \caption{Compared to the Baseline without KD. `T' and `S' denote the teacher and student model, respectively.}
        \label{tab3}
	\end{center}
\end{table}
\begin{table*}[htbp]
	\setlength{\tabcolsep}{1.6pt}
	\renewcommand{\arraystretch}{1.1}
	\begin{center}
		\begin{tabular}{l| c c c c c c c|| c c c c c c c||c c c c c c c}
			\toprule
			\textbf{Datasets} & \multicolumn{7}{c||}{\textbf{S2S-EU}} & \multicolumn{7}{c||}{\textbf{S2S-CN}} & \multicolumn{7}{c}{\textbf{M2S-GL}}\\
			\midrule
			\textbf{Methods} &\textbf{R/R} & \textbf{M/M} & \textbf{S/S} & \textbf{R/M} & \textbf{R/S} & \textbf{M/S} & \textbf{Avg.} &\textbf{R/R} & \textbf{M/M} & \textbf{S/S} & \textbf{R/M} & \textbf{R/S} & \textbf{M/S} & \textbf{Avg.} &\textbf{R/R} & \textbf{M/M} & \textbf{S/S} & \textbf{R/M} & \textbf{R/S} & \textbf{M/S} & \textbf{Avg.}\\
			\midrule
			RKD & 91.6 & 89.1 & 85.9 & 89.9 & 85.1 & 86.0 & 87.9 &94.9 & 91.9 & 90.0 & 91.7 & 90.0 & 89.7 & 91.4 & 95.1 & \underline{93.4} & 88.3 & \underline{93.6} & 88.9 & \underline{89.1} & 91.4\\
			DKD & 91.7 & \textbf{91.7} & \underline{87.1} & 90.4 & 86.2 & \underline{86.8} & \underline{89.0} & 94.9 & 93.1 & 89.8 & 92.2 & 89.5 & 90.7 & 91.7 & 68.1 & 63.6 & 64.9 & 63.5 & 59.1 & 60.0 & 63.2\\
			Logits & 87.8 & 89.1 & 73.8 & 88.8 & 84.0 & 85.9 & 84.9 & 94.7 & 92.6 & 90.0 & 91.6 & 90.5 & 90.7 & 91.7 & 93.6 & 91.2 & 85.0 & 91.4 & 86.1 & 88.1 & 89.2\\
			CTKD & 92.5 & 90.9 & 71.8 & \textbf{92.1} & \underline{86.6} & 86.0 & 86.7 & 94.8 & \underline{93.5} & \underline{90.4} & 92.2 & 90.1 & 91.1 & \underline{92.0} & 89.0 & 87.1 & 82.1 & 89.0 & 82.3 & 82.0 & 85.3\\
			LSKD & 92.1 & 88.8 & 86.8 & 89.9 & 85.5 & 85.8 & 88.2 & 95.4 & 91.9 & 89.6 & 91.4 & \textbf{91.0} & 90.8 & 91.7 & \underline{95.4} & 93.3 & \underline{90.2} & 93.1 & \underline{89.3} & \underline{89.1} & \underline{91.7}\\
            VPR & 46.2 & 75.2 & 33.9 & 69.2 & 38.8 & 40.1 & 50.6 & 94.4 & 90.0 & 88.5 & 90.1 & 88.4 & 88.7 & 90.0 & 94.1 & 91.2 & 86.7 & 91.0 & 87.9 & 85.7 & 89.4\\
            \midrule
            $\mathcal{L}_{kd}$ & \underline{92.6} & 89.3 & 86.2 & 90.1 & 85.3 & 85.3 & 88.1 & \underline{95.6} & 91.9 & 89.8 & \underline{92.3} & 89.8 & \underline{91.2} & 91.8 & 93.6 & 91.5 & 84.6 & 90.8 & 85.5 & 84.5 & 88.4\\
			\textbf{+Ours} & \textbf{93.7} & \textbf{91.7} & \textbf{88.4} & \textbf{92.1} & \textbf{87.9} & \textbf{87.8} & \textbf{90.3} & \textbf{96.2} & \textbf{93.6} & \textbf{91.4} & \textbf{93.5} & \textbf{91.0} & \textbf{91.6} & \textbf{92.9} &
            \textbf{96.6} & \textbf{93.9} & \textbf{90.8} & \textbf{93.9} & \textbf{90.3} & \textbf{89.7} & \textbf{92.5}\\
			\bottomrule
		\end{tabular}
        \caption{Compared with SOTA methods in terms of OA. $\textbf{R}$, $\textbf{M}$ and $\textbf{S}$ indicates ResNet34, MobileNetV2 and ShuffleNetV2 respectively. $\mathcal{L}_{kd}$ is based on Vanilla KD. The best results are marked in \textbf{bold} and the second best in \underline{underline}.}
        \label{tab4}
	\end{center}
\end{table*}
\subsection{Semantic-aware Knowledge Alignment (SKA)}
To optimize the transport cost between matched samples, in this section, a transport plan $\pi$ is finalized, so we name this module as Planner as shown in Figure~\ref{fig4}. Suppose the overall transport cost of two distributions $x$ and $y$ containing $m$ and $n$ samples respectively:
\begin{equation}
\label{eq11}
\ L_{OT} = \sum_{i=1}^m \sum_{j=1}^n \pi_{ij} c(x_i, y_j).
\end {equation}
To estimate the optimal transport plan $\pi_{x \rightarrow y}$ between $x$ and $y$, we utilize Lagrangian functions with boundary regularization $\sum_{j=1}^n \pi_{ij}=1$ and entropy regularization $\epsilon H(\pi)=\sum_{i,j}\pi_{ij}log\pi_{ij}$, where $\epsilon$ is an coefficient. The details of this part can be found in Appendix A. Finally, intra-modality transport plan can be formulated as:
\begin{equation}
\label{eq12}
\ \pi_{x \rightarrow y} = softmax(\frac{c(x, y)}{\epsilon}).
\end {equation}
\begin{table}[t]
	\setlength{\tabcolsep}{5pt}
	\renewcommand{\arraystretch}{0.7}
	\begin{center}
		\begin{tabular}{l||c c||c c||c c}
			\toprule
			\multirow{2}{*}{\textbf{Method}} & \multicolumn{2}{c||}{\textbf{S2S-EU}} & \multicolumn{2}{c||}{\textbf{S2S-CN}} & \multicolumn{2}{c}{\textbf{M2S-GL}}\\
			\cmidrule(r){2-3} \cmidrule(r){4-5} \cmidrule(r){6-7}
			&\textbf{OA} & \textbf{F1} &\textbf{OA} & \textbf{F1} &\textbf{OA} & \textbf{F1}\\
			\midrule
			Vanilla KD & 92.6 & 92.3 & 95.6 & 95.0 & 93.6 & 91.6\\
			+ SemBridge & \textbf{93.7} & \textbf{93.6} & \textbf{96.2} & \textbf{95.8} & \textbf{96.6} & \textbf{95.1}\\
			\midrule
			RKD & 91.6 & 91.5 & 94.9 & 94.3 & 95.1 & 93.2\\
			+ SemBridge & \textbf{92.3} & \textbf{92.2} & \textbf{95.7} & \textbf{95.3} & \textbf{95.4} & \textbf{93.6}\\
			\midrule
			DKD & 91.7 & 91.5 & 94.9 & 94.3 & 68.1 & 73.5\\
			+ SemBridge & \textbf{92.4} & \textbf{92.2} & \textbf{95.3} & \textbf{95.0} & \textbf{83.0} & \textbf{82.9}\\
			\midrule
			Logits & 87.8 & 87.5 & 94.7 & 94.2 & 93.6 & 91.2\\
			+ SemBridge & \textbf{91.4} & \textbf{91.2} & \textbf{96.0} & \textbf{95.5} & \textbf{94.3} & \textbf{92.4}\\
			\midrule
			CTKD & 92.5 & 92.3 & 94.8 & 94.3 & 89.0 & 88.0\\
			+ SemBridge & \textbf{93.3} & \textbf{93.2} & \textbf{95.8} & \textbf{95.5} & \textbf{95.4} & \textbf{93.1}\\
			\midrule
			LSKD & 92.1 & 92.0 & 95.4 & 95.0 & 95.4 & 93.1\\
			+ SemBridge & \textbf{92.7} & \textbf{92.5} & \textbf{95.9} & \textbf{95.5} & \textbf{95.5} & \textbf{93.4}\\
			\bottomrule
		\end{tabular}
        \caption{Generalization capability testing.}
        \label{tab5}
	\end{center}
\end{table}
To formulate $c(\cdot)$ and coefficient $\epsilon$ w.r.t. Equ.~(\ref{eq12}), we instantiate Equ.~(\ref{eq12}) with a learnable multi-head attention structure to avoid manual choices of $c(\cdot)$ and $\epsilon$, inspired by its resemblance to the formulation of multi-head attention noted as Planner as shown in Figure~\ref{fig4}. Specifically, $z_T$ and $z_S$ are flatten with $N$ patches as $P_T$ and $P_S$ before feding into Planner. Planner project $P_T$ and $P_S$ into $Q_T$, $K_T$ and $Q_S$, $K_S$ with $H$ heads as following:
\begin{equation}
\label{eq13}
    Q_T, K_T = \text{Planner}(P_T), \quad Q_S, K_S = \text{Planner}(P_S),
\end{equation}
and compute their transport plan as $\pi_T,\pi_S \in R^{H \times N}$ based on their correlations. 
{\small
\begin{equation}
	\label{eq14}
	\ \pi_T = softmax(Q_T \cdot \frac{K_T^{\top}}{\sqrt{d}}),~~~\pi_S = softmax(Q_S \cdot \frac{K_S^{\top}}{\sqrt{d}})
\end{equation}
}
where $d$ is the feature dimension per head.
\begin{figure}[t]
	\centering
	\setlength{\belowcaptionskip}{0.cm}
	\includegraphics[width=8.5 cm]{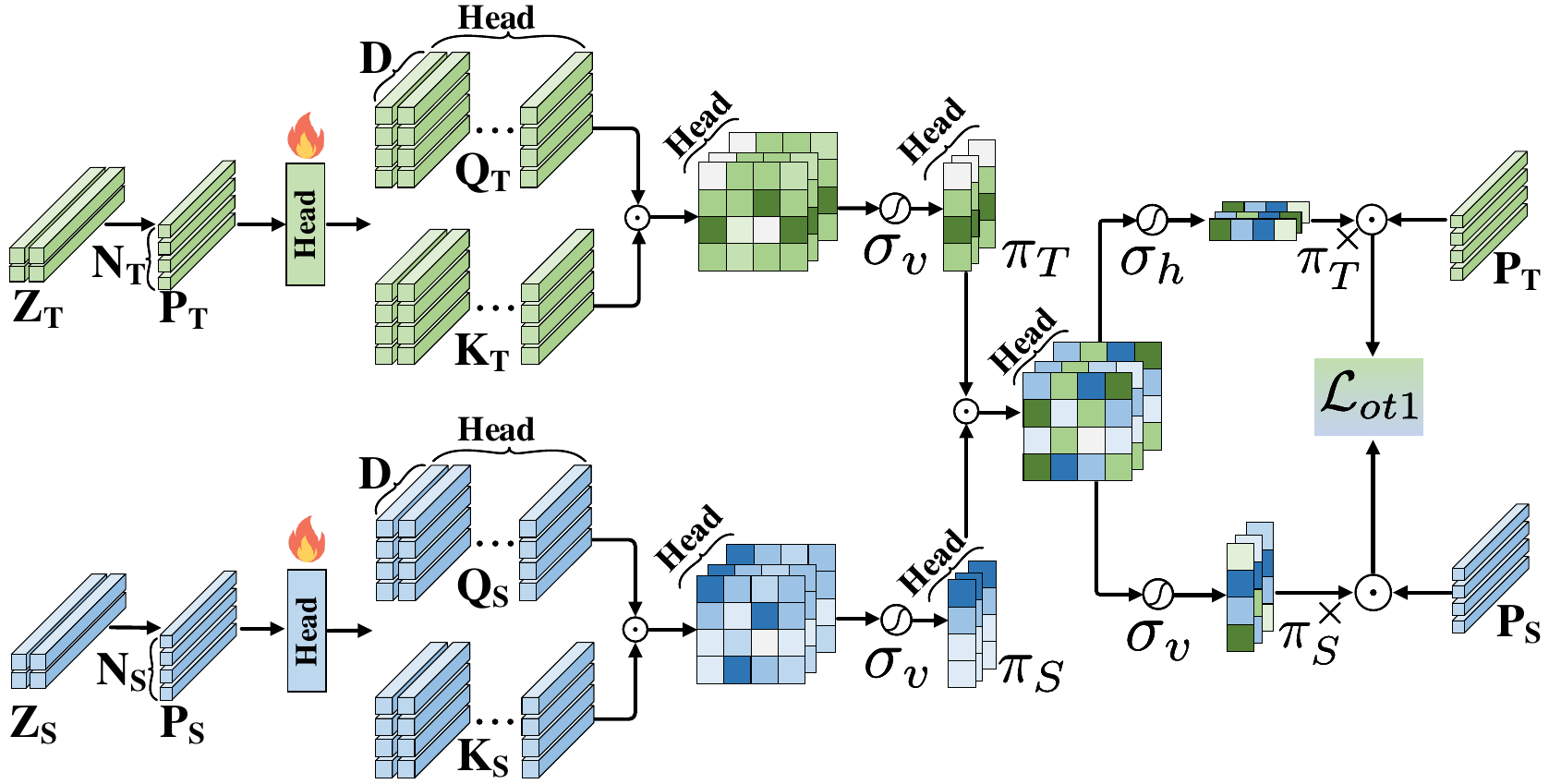}
    \caption{The structure of Planner, which is used to finalize the optimal transport cost.}
	\label{fig4}
\end{figure}
}
Subsequently, cross-modality transmission plans are implemented as:
{
\small
\begin{equation}
\label{eq15}
    \pi_T^{\times} = \frac{\sigma_h(\sigma_v(\pi_T) \cdot \sigma_v(\pi_S))}{\epsilon_T}, \quad
    \pi_S^{\times} = \frac{\sigma_v(\sigma_v(\pi_S) \cdot \sigma_v(\pi_T))}{\epsilon_S},
\end{equation}
}
where $\epsilon_{T}=\frac{1}{N_{T}} \sum_{i=1}^{N_{T}}{P_{T}}$ and $\epsilon_{S}=\frac{1}{N_{S}} \sum_{i=1}^{N_{S}}{P_{S}}$. $\epsilon_{T}$ and $\epsilon_{S}$ are scaling factors for numerical stability, computed from the average of patches $P_T$ and $P_S$. $N_{T}$ and $N_{S}$ are the number of patches of the teacher and student samples, respectively, while $\sigma_{h}$ and $\sigma_{v}$ denote the horizontal and vertical mean pooling operations. The cross-modal transmission plan is then applied to $z_T$ and $z_S$ as follows:
\begin{equation}
\label{eq16}
    D_T = z_T \cdot \frac{1}{H} \sum_{h=1}^{H} \pi_T^{\times, h}, \quad
    D_S = z_S \cdot \frac{1}{H} \sum_{h=1}^{H} \pi_S^{\times, h},
\end{equation}
Finally, to further bridge the modality gap, we employ CORAL~\cite{sun2016deep} to align refined feature $D_{T}$ and $D_{S}$ and fused feature $\overline{z}_T$ and $\overline{z}_T$ respectively and get cost $\mathcal{L}_{ot1}$ and $\mathcal{L}_{ot2}$ implemented as:
{\small
\begin{equation}
\label{eq17}
    \mathcal{L}_{ot1} = \text{CORAL}(D_T, D_S), \quad
    \mathcal{L}_{ot2} = \text{CORAL}(\overline{z}_T, \overline{z}_S).
\end{equation}
}
The overall loss function is formulated as:
\begin{equation}
	\label {eq18}
	\ \mathcal{L}_{all} = \mathcal{L}_{task} + \lambda_{1}\mathcal{L}_{kd} + \lambda_{2}(\mathcal{L}_{ot1} + \mathcal{L}_{ot2}).
\end {equation}
Here, $\mathcal{L}_{kd}\in\{$Vanilla KD~\cite{hinton2015distilling}, RKD~\cite{park2019relational}, DKD~\cite{zhao2022decoupled}, Logits~\cite{ba2014deep}, CTKD~\cite{li2023curriculum}, STKD~\cite{sun2024logit}$\}$ denotes the SCKD loss and $\mathcal{L}_{task}$ is the task-related loss. $\lambda_{1}$ and $\lambda_{2}$ are balanced factors. We set $\lambda_{2} = 1 -\lambda_{1}$. The detail of CORAL is implemented in Appendix A.

\section{Dataset Construction}
Lacking modalities with weak semantic consistency in RS scene classification tasks hampers the application of knowledge propagation. 
Therefore, a comprehensive modality paired with asymmetric information is indispensable. To address this issue, we construct a new dataset benchmark consisting of 3 sub-datasets, S2S-EU, S2S-CN, and M2S-GL with MS images and RGB images as shown in Table \ref{tab2}.
Due to unique geographical environments, scenes of the same category in RS images always present various semantic content worldwide. The goal of this research is to propagate knowledge from any place or country to others, regardless of the semantic content. 
To do this, we investigated and collected available MS images from 3 public datasets, i.e., EuroSAT~\cite{helber2018introducing}, NaSC-TG2~\cite{Zhou2021NaSC}, and BigEarthNet~\cite{sumbul2019bigearthnet}, respectively, which contain scenes from around the world. 
Subsequently, we collected RGB images from other public datasets as an asymmetric modality. The details of the proposed dataset benchmark can be found in Appendix B. 
Finally, to evaluate the difficulty of knowledge propagation in ACKD, we compute the mutual information within class on 3 proposed datasets, which is shown in Figure 6 in Appendix B.
\begin{table}[t]
\setlength{\tabcolsep}{15pt}
\renewcommand{\arraystretch}{0.8}
\centering
\begin{tabular}{lcccc}
\toprule
\textbf{$\gamma$} & 1 & 3 & 5 & 7\\
\midrule
\textbf{OA} & 93.3 & \textbf{93.7} & 93.2 & 93.1\\
\bottomrule
\end{tabular}
\caption{The impact of temperature $\gamma$ on S2S-EU. The teacher and student are both ResNet34.}
\label{tab6}
\end{table}

\section{Experiments}
\subsection{Experimental Setup}
\textbf{Datasets.} Self-constructed benchmarks involving S2S-EU, S2S-CN, and M2S-GL are employed to evaluate SemBridge for RS scene classification tasks. Specifically, S2S-EU and S2S-CN are used to evaluate the performance in single-label$\rightarrow$single-label classification, while M2S-GL is employed to assess knowledge propagation from multi-label$\rightarrow$single-label classification.

\textbf{Evaluation metrics.} Overall Accuracy (OA) and F1-score (F1) are utilized to evaluate classification performance. Following the setup in~\cite{LiuH}, only F1 is used to evaluate teacher performance on multi-label classification tasks in M2S-GL.

\textbf{Compared method.} We compared several methods in this experimental section. `Baseline' denotes the original training without KD. We also employ knowledge distillation approaches Vanilla KD~\cite{hinton2015distilling}, RKD~\cite{park2019relational}, DKD~\cite{zhao2022decoupled}, Logits~\cite{ba2014deep}, CTKD~\cite{li2023curriculum}, LSKD~\cite{sun2024logit}$\}$, and VPR~\cite{wang2024distilvpr} to evaluate the performance on ACKD compared with applying the proposed SemBridge(+SemBridge).

\textbf{Evaluated Network.} Experiment are conducted over ResNet34~\cite{he2016deep}, MobileNetV2~\cite{sandler2018mobilenetv2} and ShuffleNetV2~\cite{ma2018shufflenet}. The whole training details is implemented in Appendix C.

\subsection{Compared with Baseline Methods}
In Table~\ref{tab3}, we conduct experiments on both homogeneous and heterogeneous model architectures. Compared to baseline without KD, the SemBridge with Vanilla KD enables the student model to achieve significant improvements across 3 datasets. For example, for a homogeneous model of ResNet34, SemBridge leads to 1.3\%$\sim$2.0\% and 1.4\%$\sim$2.0\% gains on S2S-EU, S2S-CN, and M2S-GL in terms of OA and F1, respectively. 
Furthermore, to evaluate the performance of SemBridge under the different architectures between the teacher and the student, ShuffleNetV2 and MobileNetV2 are supervised by homogeneous and heterogeneous teachers, respectively. 
The results indicate that knowledge can be propagated effectively via SemBridge regardless of model architectures.

\subsection{Compared with State-of-the-art Methods}
Table~\ref{tab4} reports the classification performance of SemBridge based on Vanilla KD across 6 combinations of model architecture. SemBridge enables Vanilla KD to achieve SOTA performance on 3 datasets in terms of OA. For ResNet34, SemBridge enables Vanilla KD to achieve improvements of 0.6\% and 1.1\% on S2S-EU and S2S-CN, and outperforms LSKD by 1.2\% on M2S-GL. For ShuffleNetV2 supervised by ResNet34, our approach outperforms CTKD and LSKD with gains of 1.3\% and 1.0\% on S2S-EU and M2S-GL, respectively. Compared to uni-modality-based methods (Vanilla KD, RKD, DKD, Logits, LSKD, CTKD), VPR is designed to distill knowledge between modalities with the same semantic content. It can be found that due to semantic differences, VPR shows unpromising results, especially on S2S-EU. It also indicates the necessity of ACKD.
\begin{table}[t]
	\setlength{\tabcolsep}{3.5pt}
	\renewcommand{\arraystretch}{1.0}
	\begin{center}
		\begin{tabular}{c c c c |c c c}
			\toprule
			SSM & DynM & $\mathcal{L}_{ot1}$ & $\mathcal{L}_{ot2}$& \textbf{S2S-EU} & \textbf{S2S-CN} & \textbf{M2S-GL}\\
			\midrule
			\textcolor{red}{\ding{55}} & \ding{51} & \ding{51} & \ding{51} & 92.5 & 95.3 & 95.6\\
			\ding{51} & \textcolor{red}{\ding{55}}  & \ding{51} & \ding{51} & 92.9 & 95.1 & 94.2\\
			\ding{51} & \ding{51} & \textcolor{red}{\ding{55}} & \ding{51} & 92.5 & 96.1 & 95.1\\
			\ding{51} & \ding{51} & \ding{51} & \textcolor{red}{\ding{55}} & 92.8 & 94.1 & 95.8\\
			\ding{51} & \ding{51} & \ding{51} & \ding{51} & \textbf{93.7} & \textbf{96.2} & \textbf{96.6}\\
			\bottomrule
		\end{tabular}
        \caption{Impact of SSM, DynM, and SKA($\mathcal{L}_{ot1}$, $\mathcal{L}_{ot2}$) of SemBridge on R/R in terms of OA.}
        \label{tab7}
	\end{center}
\end{table}

\subsection{Generalization Capability Testing}
Table~\ref{tab5} illustrates the generalization capability testing on ResNet34. It can be observed that SemBridge can enhance the performance of existing SCKD approaches on ACKD tasks. Compared to others, SemBridge with Vanilla KD achieves the best performance with 93.7\%$\sim$96.6\% and 93.6\%$\sim$95.8\% in terms of OA and F1, respectively. On single-label$\rightarrow$single-label tasks, SemBridge shows the greatest improvement based on Logits with gains of 3.6\% and 3.7\%, and 1.3\% and 1.3\% in terms of OA and F1, respectively. On multi-label$\rightarrow$single-label tasks, SemBridge achieves the largest improvement for DKD, with increases of 14.9\% and 9.4\% of OA and F1, respectively. It should be noted that we only applied SemBridge to uni-modality-based methods, which are typically used on more universal scenarios.
\begin{figure}[t]
	\centering
	\setlength{\belowcaptionskip}{0.cm}
	\includegraphics[width=8.5cm]{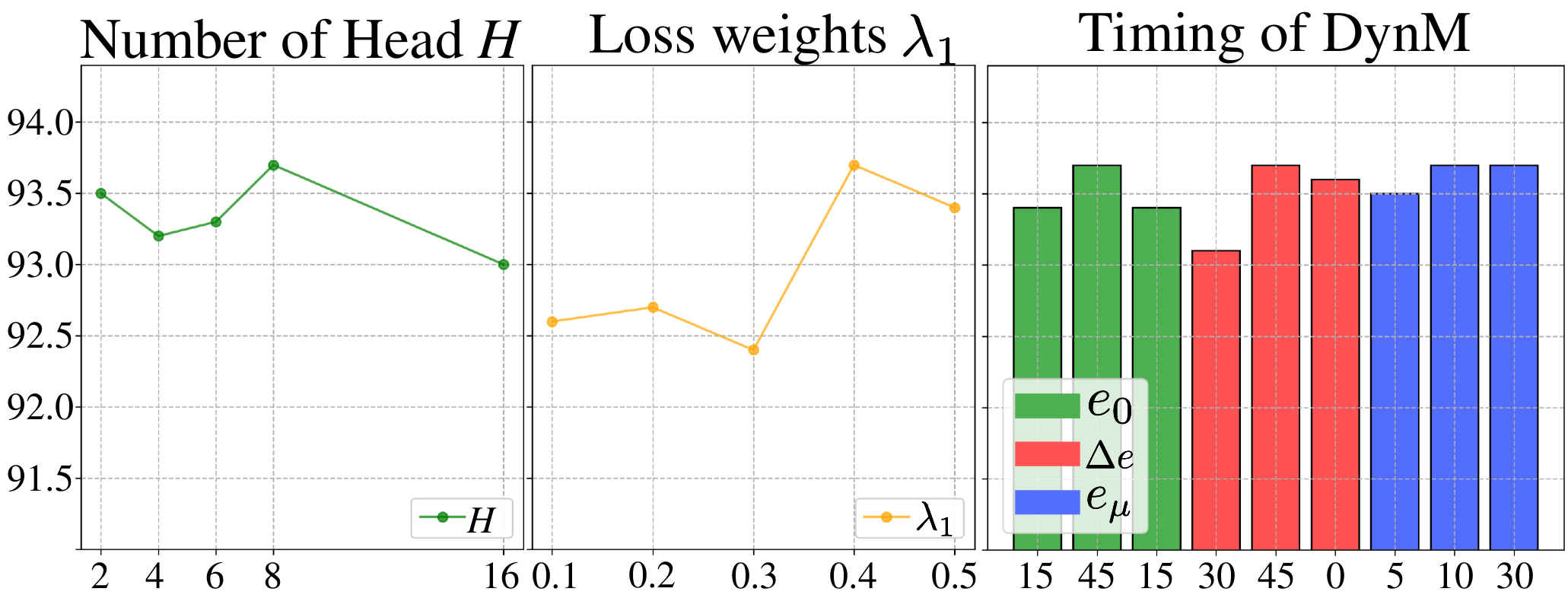}
    \caption{Hyperparameter analysis.}
	\label{fig5}
\end{figure}

\begin{table}[t]
\centering
\setlength{\tabcolsep}{11pt}
\renewcommand{\arraystretch}{1.0}
\begin{tabular}{lccc}
\toprule
\textbf{Dataset} & \textbf{S2S-EU} & \textbf{S2S-CN} & \textbf{M2S-GL} \\
\midrule
DynM & 5.6 & 2.9 & 4.5 \\
Planner & 2.0 & 1.6 & 7.4 \\
\midrule
Vanilla KD & 64.1 & 51.5 & 63.9 \\
\textbf{$\Delta$ (\%)} & +11.9 & +8.7 & +18.6 \\
\bottomrule
\end{tabular}
\caption{Training Speed Analysis (mins). $\Delta$=(DynM + Planner) / baseline $*$ 100\% where baseline indicates Vanilla KD.}
\label{tab8}
\end{table}

\subsection{Hyperparameter Anaylsis and Ablation Study}
As shown in Table~\ref{tab6}, we found that SemBridge achieves the best OA at 93.7\% when $\gamma=3$ in DynM.
As shown in Figure~\ref{fig5}, we analyze the effects of the number of heads $H$ in Planner and the loss weight $\lambda_1$. The optimal performance is observed when $H = 8$ and $\lambda_1 = 0.4$. Furthermore, the timing of DynM, i.e., $e_0$, $\Delta e$, and $e_{\mu}$ exhibit consistent robustness across settings. Besides, we investigate the effectiveness of each component in SemBridge as shown in Table~\ref{tab7}. All four components contribute to the best result, which is 93.7\% on S2S-EU, 96.2\% on S2S-CN, and 96.6\% on M2S-GL, indicating that SemBridge can optimize the cost caused by weak semantic consistency. The impact of DynM and the Planner on optimal transport can be found in Appendix C.

\section{Conclusion}
In this paper, we propose ACKD, a new research direction to broaden the application scope of SCKD. To this end, we construct a dataset benchmark comprising 3 sub-datasets in the remote sensing fields. Subsequently, we propose a framework, namely SemBridge, consisting of a Student-Friendly Matching module and a Semantic-Aware Knowledge Alignment module to reduce the transport cost during knowledge distillation. The experimental results demonstrate that the proposed SemBridge not only helps Vanilla KD achieve state-of-the-art performance across various datasets but also enhances the performance of existing SCKD methods on ACKD, indicating superior generalization capability. However, we also identify some limitations of SemBridge. The time consumption associated with student-friendly matching may negatively impact training speed, as shown in Table~\ref{tab8}. We regard it as the future direction.

\begin{figure*}[t]
	\centering
	\includegraphics[width=17.5cm]{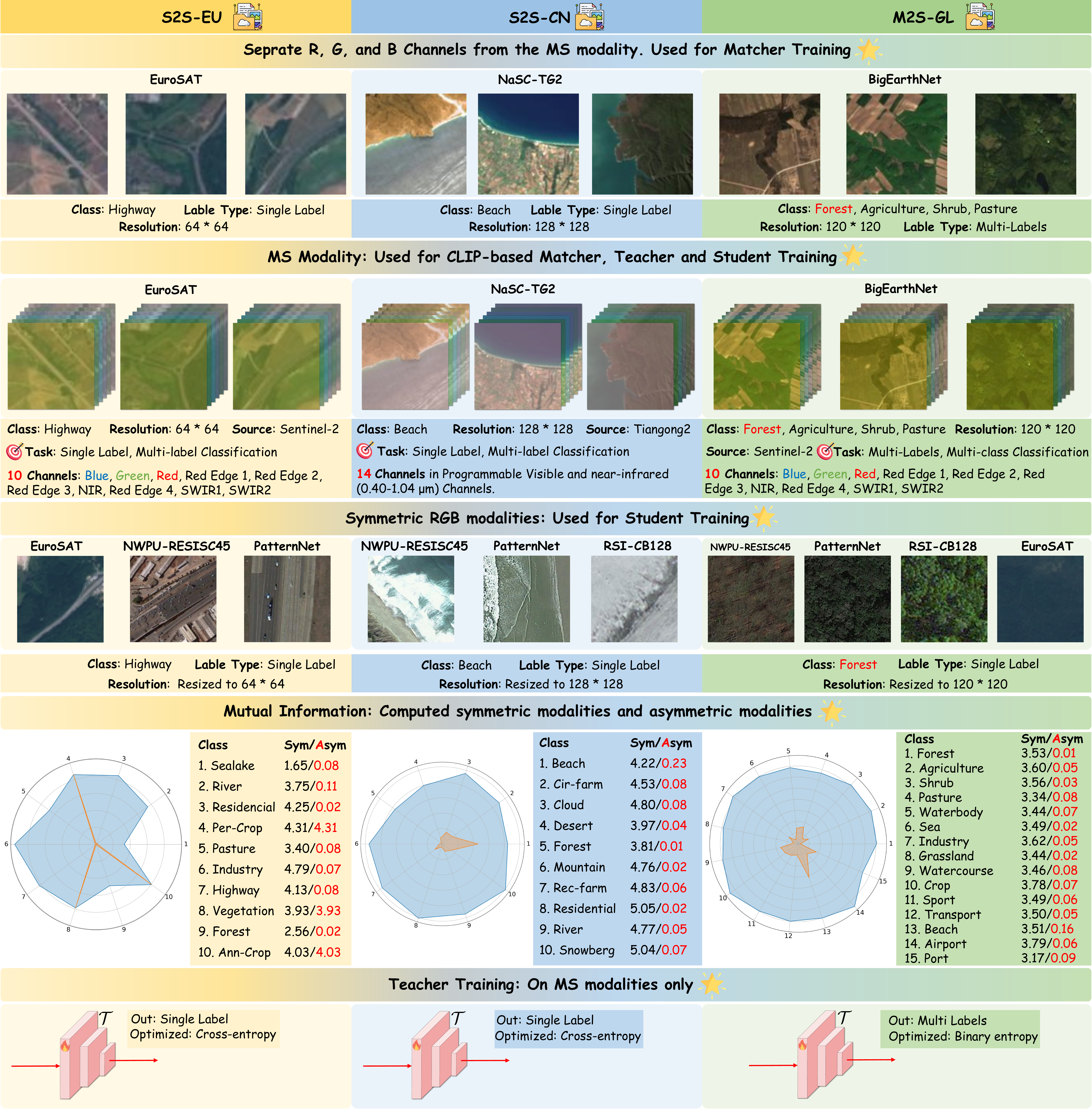}
	\caption{Illustration of the proposed dataset benchmark. This benchmark consists of 3 sub-datasets namely S2S-EU, S2S-CN, and M2S-GL respectively. On each sub-dataset, MS modality and unpaired RGB modality are collected by various equipment from different regions.}
	\label{fig6}
\end{figure*}

\begin{figure*}[t]
	\centering
	\includegraphics[width=17.5cm]{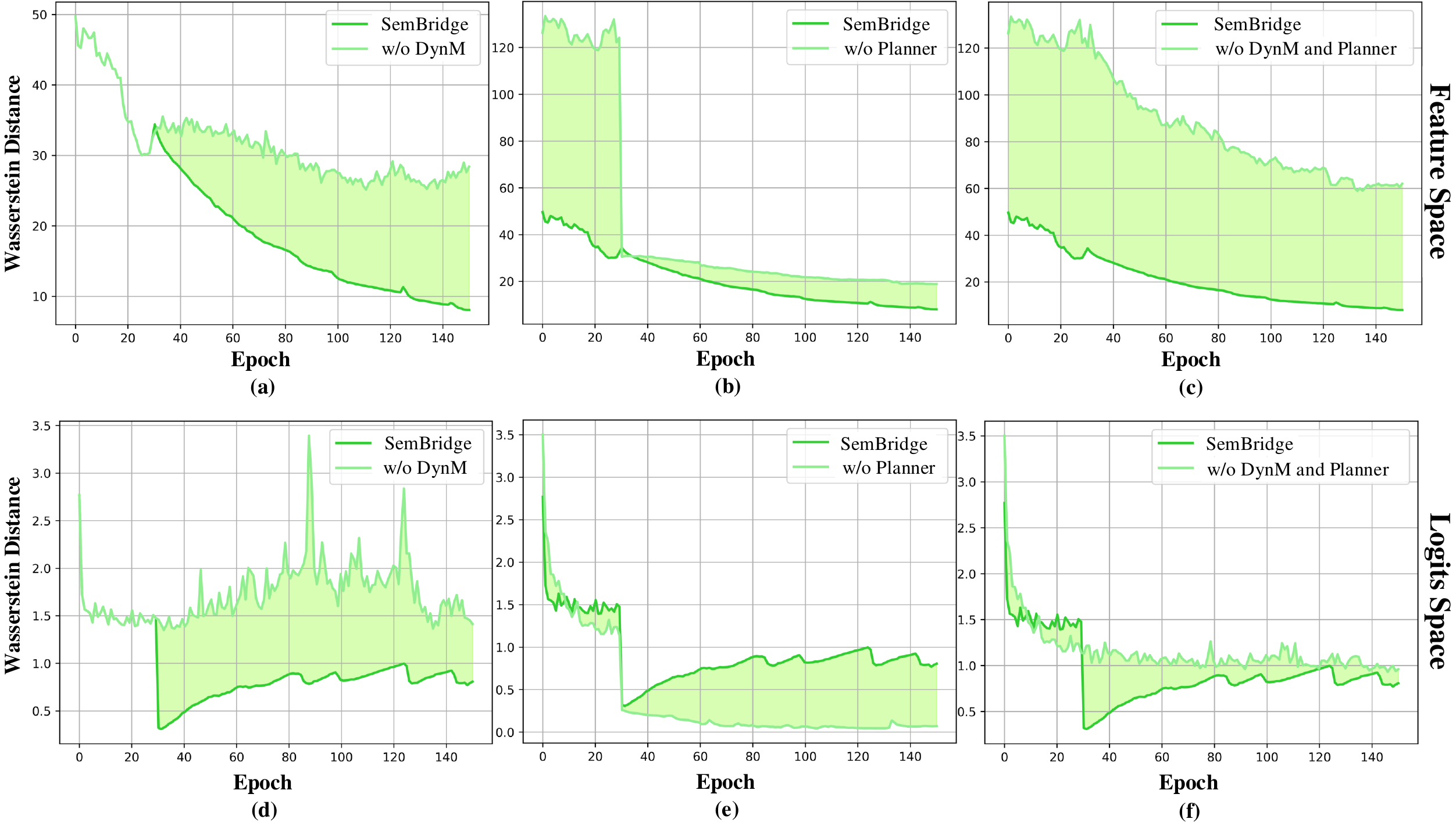}
	\caption{Impact of DynM and the Planner on optimal transport cost in feature and logits space.}
	\label{fig7}
\end{figure*}

\section{A. Methodology}

\subsection{A1. Intra-modal Transport Plan}
The cost of transport with boundary normalization item can be written as:
\begin{equation}
\label{eq22}
\ L_{OT} = \sum_{i=1}^m \sum_{j=1}^n \pi_{ij} c(x_i, y_j),  s.t. \sum_{j=1}^n \pi_{ij}=1.
\end {equation}
We further introduce Entropy Regularization $H(\pi)=\sum_{i,j}\pi_{ij}log\pi_{ij}$. Based on this, the Lagrangian optimization is introduced and reformulate Eq.~(\ref{eq8}): as:
{
\small
\begin{equation}
\label{eq23}
\ L_{OT} = \sum_{i=1}^m \sum_{j=1}^n \pi_{ij} c(x_i, y_j) - \sum_{i}{\alpha_{i}}{\sum_{j}(\pi_{ij}-1)} - \epsilon \sum_{i,j}\pi_{ij}log\pi_{ij}.
\end{equation}
}
Then, the derivation of $\pi_{ij}$ is estimated:
\begin{equation}
\label{eq24}
\ \frac{\partial {L_{OT}}}{\partial \pi_{ij}} = c(x_i, y_j) - \alpha_i - \epsilon(1+log\pi_{ij})= 0.
\end {equation}
Then, we obtain $\pi_{ij}$ as:
\begin{equation}
\label{eq25}
\ \pi_{ij} = e^{(-\frac{\epsilon + \alpha_{i} - c(x_i, y_j)}{\epsilon})}.
\end {equation}

After normalization, $\pi_{ij}$ can be written as: 
\begin{equation}
\label{eq26}
\begin{aligned}
\ \pi_{ij} &= \frac{e^{\frac{c(x_i, y_j)}{\epsilon}}}{\sum_{j'} e^{\frac{c(x_i, y_j^{'})}{\epsilon}}}.
\end{aligned}
\end {equation}

Hence, the transport plan $\pi_{i\rightarrow y}$ from single sample $x_i$ to the distribution $y$ is refined as:
\begin{equation}
\label{eq27}
\ \pi_{i\rightarrow y} = softmax(\frac{c(x_i, y)}{\epsilon}).
\end {equation}

As such, the transport plan $\pi_{x \rightarrow y}$ between $x$ and $y$ is formulated:
\begin{equation}
\label{eq28}
\ \pi_{x \rightarrow y} = softmax(\frac{c(x, y)}{\epsilon}).
\end {equation}

\subsection{A2. Implementation of CORAL}
CORAL is utilized in the semantic aware knowledge alignment module to bridge the modality gap between the teacher and the student, which is implemented as:
\begin{equation}
	\label {eq29}
	\ \mathcal{L}_{coral} = \frac{1}{4d_c} {||C_{s} - C_{t}||}_{F}^{2},
\end {equation}
where ${||\cdot||}_{F}^{2}$ is the squared matrix Frobenius norm, $d_c$ denotes the feature dimension and $C_{s}$ and $C_{t}$ are defined as follows:

\begin{equation}
	\label {eq30}
	\ C_{S} = \frac{1}{\mathcal{B}-1}(R_{S}^{\top} \cdot R_{S} - \frac{1}{\mathcal{B}}(1^{\top}R_{S})^{\top}(1^{\top}R_{S})),
\end {equation}
\begin{equation}
	\label {eq31}
	\ C_{T} = \frac{1}{\mathcal{B}-1}(R_{T}^{\top} \cdot R_{T} - \frac{1}{\mathcal{B}}(1^{\top}R_{T})^{\top}(1^{\top}R_{T})).
\end {equation}
CORAL is applied to unfused- and fused-feature alignment respectively. In the former case, $R_{T}$ and $R_{S}$ indicate enhanced representation $D_{T}$ and $D_{S}$ respectively while $R_{T}$ and $R_{S}$ represents $\overline{z}_{T}$ and $\overline{z}_{S}$ in the latter case. $\mathcal{B}$ is the batch size.

\section{B. Dataset Construction}
We construct a new dataset benchmark comprising three sub-datasets, featuring MS images and RGB images, to investigate knowledge propagation between modalities with different semantics, as shown in Figure \ref{fig6}.

\textbf{S2S-EU} aims at investigate the effectiveness of knowledge distillation from \textbf{S}ingle label MS images to \textbf{S}ingle label RGB images. To do this, MS images with 10 spectral bands obtained via Sentinel-2 are carefully cleaned. For the corresponding RGB image, we collect images from the EuroSAT, NWPU-RESISC45~\cite{ChengGong2}, and PatternNet~\cite{zhou2018patternnet} datasets. S2S-EU contains 10 classes, which are industrial Buildings, Residential Buildings, Annual Crop, Permanent Crop, River, Sea/Lake, Herbaceous Vegetation, Highway, Pasture, and Forest, respectively. Both MS and RGB images are resized to $64\times64$ pixels.

\textbf{S2S-CN.} Due to the difference in spectral bands caused by different collecting devices, MS images might show varying performances. To investigate the robustness of our framework and demonstrate that knowledge can be distilled without reliance on specific devices, we collected MS images from NaSC-TG2, which were captured by Tiangong2, the first space laboratory in China. Unlike Sentinel-2, Tiangong-2 provides 14 spectral bands. According to the 10 classes of NaSC-TG2, we collected RGB images within those classes from the NWPU-RESISC45, PatternNet, and RSI-CB128~\cite{li2020RSI-CB} datasets containing MS and RGB images resized to $128\times128$ with a single label.

\textbf{M2S-GL.} Due to the broad field of view of satellites, RS images always encompass multiple types of scenes, resulting in multiple labels. In this subset, we investigate knowledge propagation from \textbf{M}ulti-label to \textbf{S}ingle-label tasks between asymmetric modalities. Firstly, we collect multi-label MS images from several common classes from BigEarthNet, which were captured by Sentinel-2 with 10 spectral bands. RGB images with a single label are collected from NWPU-RESISC45, PatternNet, RSI-CB128, and EuroSAT datasets, respectively. After careful cleaning, we retain only 15 classes: forest, agriculture, shrub, pasture, waterbody, sea, industry, grassland, watercourse, crop, sport, transport, beach, airport, and port. For resolution, both MS and RGB images are resized to $120\times120$ pixels.

\textbf{Mutual Information Visualization}. As shown in Figure~\ref{fig6}, the mutual information of symmetric and asymmetric modality pairs of each class in three datasets is evaluated. Results indicate that symmetric modalities with the same semantic content exhibit much higher mutual information than asymmetric pairs in our benchmark. According to prior research~\cite{ahn2019variational}, higher mutual information correlates with efficient knowledge distillation. Thus, conducting knowledge from our benchmark is notably more challenging. For each class $C$, we separate the R, G, B channels from MS data as $A$, and use the RGB image as $B$. Mutual information is implemented as:
\begin{equation}
	\label {eq32}
    \ MI(A, B) = H(A) + H(B) - H(A, B),
\end{equation}
where $H(A)$ and $H(A)$ are information entropy of image A and B. $H(A, B)$ are the joint entropy. Information entropy and joint entropy are implemented as:
\begin{equation}
	\label {eq33}
    \ H(A) = - \sum_{i=0}^{N-1}{p_{i}logp_{i}}
\end{equation}
\begin{equation}
	\label {eq34}
    \ H(A,B) = - \sum_{i,j}{p_{AB}(i, j)logp_{AB}(i,j)}
\end{equation}
Here, $N$ denotes the number of pixel values equal to 256, $p_{i}$ is the probability of pixel value $i$, and $p_{AB}(i, j)$ indicates the likelihood that a pixel has a value $i$ in $A$ and $j$ in $B$ at the same spatial location. The final score is the average MI across all samples in class $C$.

\begin{table}[t]
\setlength{\tabcolsep}{15pt}
\centering
\caption{The impact of $\gamma$ on S2S-EU. The teacher and student are both ResNet34.}
\label{tabv5x6}
\begin{tabular}{lcccc}
\toprule
\textbf{$\gamma$} & 1 & 3 & 5 & 7\\
\midrule
\textbf{OA} & 93.3 & \textbf{93.7} & 93.2 & 93.1\\
\bottomrule
\end{tabular}
		\label{tab9}
\end{table}

\begin{table*}[t]
    \setlength{\tabcolsep}{20pt}
    \renewcommand{\arraystretch}{1.6}
    \begin{center}
        \caption{The details of the proposed dataset benchmark.}
        \label{tab10}
        \begin{tabular}{c c c c c c c}
            \hline
            \textbf{Class} & \multicolumn{2}{c}{\textbf{S2S-EU}} & \multicolumn{2}{c}{\textbf{S2S-CN}} & \multicolumn{2}{c}{\textbf{M2S-GL}} \\
            \cline{2-7}
            & \textbf{MS} & \textbf{RGB} & \textbf{MS} & \textbf{RGB} & \textbf{MS} & \textbf{RGB} \\
            \hline
            Sealake & 3000 & 3000 & - & - & - & - \\
            River & 2500 & 2499 & 2000 & 2000 & - & - \\
            Residencial & 3000 & 2800 & - & - & - & - \\
            Per-Crop & 2500 & 2500 & - & - & - & - \\
            Pasture & 2000 & 2000 & - & - & 2000 & 2000 \\
            Industry & 2500 & 2500 & - & - & 2000 & 2000 \\
            Highway & 2500 & 2500 & - & - & - & - \\
            Vegetation & 3000 & 3000 & - & - & - & - \\
            Forest & 3000 & 2998 & 2000 & 1500 & 2000 & 2000 \\
            Ann-Crop & 3000 & 3000 & - & - & - & - \\
            Beach & - & - & 2000 & 2000 & 822 & 2000 \\
            Cir-farm & - & - & 2000 & 700 & - & - \\
            Cloud & - & - & 2000 & 700 & - & - \\
            Desert & - & - & 2000 & 2000 & - & - \\
            Mountain & - & - & 2000 & 1664 & - & - \\
            Rec-farm & - & - & 2000 & 1228 & - & - \\
            Residential & - & - & 2000 & 2000 & - & - \\
            Snowberg & - & - & 2000 & 1667 & - & - \\
            Agriculture & - & - & - & - & 2000 & 2000 \\
            Shrub & - & - & - & - & 2000 & 1451 \\
            Waterbody & - & - & - & - & 2000 & 700 \\
            Sea & - & - & - & - & 2000 & 1494 \\
            Grassland & - & - & - & - & 2000 & 1977 \\
            Watercourse & - & - & - & - & 2000 & 1971 \\
            Crop & - & - & - & - & 2000 & 2000 \\
            Sport & - & - & - & - & 2000 & 2000 \\
            Transport & - & - & - & - & 1418 & 2000 \\
            Airport & - & - & - & - & 517 & 700 \\
            Port & - & - & - & - & 147 & 2000 \\
            \hline
        \end{tabular}
    \end{center}
\end{table*}

\section{C. Experiments}
\subsection{C1. Training details.}
The Adam optimizer is employed with a learning rate of 0.001, training on 1 NVIDIA 2080Ti GPU over 200 epochs. The batch size is set to 128. In the matcher training stage, we follow the configuration of the original CLIP~\cite{radford2021learning} and $\mathcal{M}_{V}$ and $\mathcal{M}_{G}$ employ ResNet34-based architecture~\cite{he2016deep}. In DynM, $\gamma$ is set to 3. In the teacher training stage, cross-entropy loss is applied for single-label classification tasks in S2S-EU and S2S-CN, while binary-entropy loss is used for multi-label classification tasks in M2S-GL.

\subsection{C2. Impact of DynM and Planner on transport cost}
Dynamic Matching (DynM) and Planner play important roles in the learning journey of the student. The former enables the student to seek knowledge from different teachers according to their current capability. The latter plans an optimal transport for their knowledge propagation to reduce the cost. To understand their contribution, we visualize the Wasserstein distance without (w/o) those components and compare them to the whole framework in feature and logits space as shown in Figure~\ref{fig7}. In feature space, when applying both DynM and the Planner, the optimization cost is decreased more sharply. This indicates the effectiveness of the DynM and the Planner. In logits space, despite slightly increased cost caused by Planner, utilizing DynM and the Planner together can also finalize the transport cost.

\begin{algorithm*}[!h]
	\caption{Algorithm of SemBridge.}
	\label{alg:AOS}
	\renewcommand{\algorithmicrequire}{\textbf{Input:}}
	\renewcommand{\algorithmicensure}{\textbf{Initialize:}}
	\renewcommand{\algorithmiccomment}[1]{\hfill $\triangleright$ #1}
	\begin{algorithmic}[1]
		\REQUIRE $\mathcal{D}\{V_{K_c}^{c},G_{N_c}^{c}\}$  
		\ENSURE $\mathcal{T}(\theta_T)$, $\mathcal{S}(\theta_S)$, $\mathcal{M}(\theta_\mathcal{M})$, $\mathcal{M}_V(\theta_{\mathcal{M}_V})$, $\mathcal{M}_G(\theta_{\mathcal{M}_G})$, $Planner(\theta_{P})$  
		\ENSURE learning rate $\alpha$, cross-entropy loss $\mathcal{L}_{CE}$, binary-entropy loss $\mathcal{L}_{BCE}$
		\ENSURE $e_t = e_0 + \sum_{i=1}^{t} (\Delta e + e_{\mu} (i-1))$
		\STATE \textbf{$\#$ Training Teacher}
		\FOR{epoch in iterations}
		\STATE\textbf{$\#$ Single-label Classification}
		\STATE $\theta_\mathcal{T} \gets \theta_\mathcal{T} - \alpha \bigtriangledown_{\theta_\mathcal{T}}\mathcal{L}_{CE}$
		\STATE\textbf{$\#$ Multi-label Classification}
		\STATE $\theta_\mathcal{T} \gets \theta_\mathcal{T} - \alpha \bigtriangledown_{\theta_\mathcal{T}}\mathcal{L}_{BCE}$
		\ENDFOR
		\STATE \textbf{$\#$ Training semantic aware matcher}
		\FOR{epoch in iterations}
            \STATE $\tilde{G} \gets channel_{split}(V)$
		\STATE  $\mathcal{L}_{semantic} = InfoNCE(\mathcal{M}_V(V),\mathcal{M}_G(\tilde{G}))$
		\STATE $\theta_\mathcal{M} \gets \theta_\mathcal{M} - \alpha \bigtriangledown_{\theta_\mathcal{M}}\mathcal{L}_{semantic}$
		\ENDFOR
            \STATE \textbf{$\#$ Construct teacher galleries $\mathcal{G}_{1}$ and $\mathcal{G}_{2}$}
            \FOR{$c$ in $C^{th}$ classes}
            \FOR{$V_{k}^{c}$ in $V^{c} \in \mathcal{D}$}
            \STATE $v_k^c = \mathcal{M}_V(V_k^c)$, $\textbf{p}_{T,k}^c = \mathcal{T}(V_k^c)$
            \STATE $\mathcal{G}_{1} \gets v_k^c$, $\mathcal{G}_{2} \gets \textbf{p}_{T,k}^c$
            \ENDFOR
            \ENDFOR
		\STATE \textbf{$\#$ Initialization of $\mathcal{D}_{match}\{V_{N_c}^{c},G_{N_c}^{c}\}$}
		\FOR{$c$ in $C^{th}$ classes}
		\FOR{$G_{n}^{c}$ in $G^{c} \in \mathcal{D}$}
		\STATE  $g_n^c \gets \mathcal{M}_G(G_n^c)$
		\FOR{$v_{k}^{c} \in \mathcal{G}_{1}$}
		\STATE  $v_k^c \gets \mathcal{M}_V(V_k^c)$
		\STATE $\Phi_{n}^{c} \gets \textbf{cos}(g_n^c, v_k^c)$
		\ENDFOR
		\STATE $\textbf{k}^* = \arg\max_{\textbf{k}} \Phi_{n,\textbf{k}}^c$
		\STATE \textbf{Update} $\mathcal{D}_{\text{match}}^c = \left\{ \left( V_{\textbf{k}^*}^c, G_n^c \right) \mid n = 1, \dots, N_c \right\}$
		\ENDFOR
		\ENDFOR
		\STATE \textbf{Output:} $\mathcal{D}_{match}\{V_{N_c}^c,G_{N_c}^c\}$
		\STATE \textbf{$\#$ Student Training on $\mathcal{D}_{match}$}
		\FOR{epoch in iterations}
		\STATE  $\theta_\mathcal{S} \gets \theta_\mathcal{S} - \alpha \bigtriangledown_{\theta_\mathcal{S}}\mathcal{L}_{all}$
		\STATE  $\theta_{P} \gets \theta_{P} - \alpha \bigtriangledown_{\theta_{P}}\mathcal{L}_{all}$
		\STATE \textbf{$\#$Update $\mathcal{D}_{match}\{V_{N_c}^{c},G_{N_c}^{c}\}$}
		\IF {epoch $\in e_t$}
		
		\FOR{$c$ in $C^{th}$ classes}
		\FOR{$G_{n}^{c}$ in $G^{c} \in \mathcal{D}$}
		\STATE  $\textbf{p}_{S,n}^{c} \gets \mathcal{S}(G_n)$
		
		\FOR{$\textbf{p}_{T,k}^{c} \in \mathcal{G}_{2}$}
		\STATE $\Omega_{n,k}^{i} \gets \mathrm{KL}(\textbf{p}_{S,n}^c \| \textbf{p}_{T,k}^{c})$
		\ENDFOR
        \STATE $\Omega_n^c=\left[ 
        \mathrm{KL}(\textbf{p}_{S,n}^c \| \textbf{p}_{T,1}^c),\ 
        \mathrm{KL}(\textbf{p}_{S,n}^c \| \textbf{p}_{T,2}^c),\ 
        \dots,\ 
        \mathrm{KL}(\textbf{p}_{S,n}^c \| \textbf{}_{T, K_c}^c)
    \right].$
		\STATE $\textbf{k}^* = \arg\min_\textbf{k} \Omega_{n,\textbf{k}}^c$
		\STATE \textbf{Update} $\mathcal{D}_{\text{match}}^c = \left\{ \left( V_{\textbf{k}^*}^c, G_n^c \right) \mid n = 1, \dots, N_c \right\}$
		\ENDFOR
		\ENDFOR
		
		\ENDIF
		\ENDFOR
		
	\end{algorithmic}
\end{algorithm*}

\clearpage
\bibliography{aaai2026}

\end{document}